\documentclass[runningheads,a4paper]{llncs}

\usepackage{amsmath}
\usepackage{amssymb}
\usepackage[misc]{ifsym}
\usepackage{graphicx}
\usepackage{color}

\usepackage{algorithm}
\usepackage{algorithmicx}
\usepackage{algpseudocode}
\usepackage{multirow}
\usepackage{array}
\newcolumntype{L}[1]{>{\raggedright\let\newline\\\arraybackslash\hspace{0pt}}m{#1}}
\newcolumntype{C}[1]{>{\centering\let\newline\\\arraybackslash\hspace{0pt}}m{#1}}
\newcolumntype{R}[1]{>{\raggedleft\let\newline\\\arraybackslash\hspace{0pt}}m{#1}}

\usepackage{url}
\urldef{\mailsa}\path|{alfred.hofmann, ursula.barth, ingrid.haas, frank.holzwarth,|
\urldef{\mailsb}\path|anna.kramer, leonie.kunz, christine.reiss, nicole.sator,|
\urldef{\mailsc}\path|erika.siebert-cole, peter.strasser, lncs}@springer.com|

\begin{document}

\mainmatter  

\title{A Fixed-Point Model for \\ Pancreas Segmentation in Abdominal CT Scans}

\titlerunning{A Fixed-Point Model for Pancreas Segmentation in Abdominal CT Scans}

%
%
\author{Yuyin Zhou\textsuperscript{1}, Lingxi Xie\textsuperscript{2}$^{(\textrm{\Letter})}$, Wei Shen\textsuperscript{3},\\
Yan Wang\textsuperscript{4}, Elliot K. Fishman\textsuperscript{5}, Alan L. Yuille\textsuperscript{6}}
\authorrunning{Y. Zhou {\em et al.}}

\institute{
\textsuperscript{1,2,3,4,6}The Johns Hopkins University, Baltimore, MD 21218, USA\\
\textsuperscript{3}Shanghai University, Baoshan District, Shanghai 200444, China\\
\textsuperscript{5}The Johns Hopkins University School of Medicine, Baltimore, MD 21287, USA\\
\textsuperscript{1}{\tt\small zhouyuyiner@gmail.com}\quad
\textsuperscript{2}{\tt\small 198808xc@gmail.com}\quad
\textsuperscript{3}{\tt\small wei.shen@t.shu.edu.cn}\\
\textsuperscript{4}{\tt\small wyanny.9@gmail.com}\quad
\textsuperscript{5}{\tt\small efishman@jhmi.edu}\quad
\textsuperscript{6}{\tt\small alan.l.yuille@gmail.com}\\
\textcolor{red}{{\tt http://bigml.cs.tsinghua.edu.cn/\~{}lingxi/Projects/OrganSegC2F.html}}
}

%
%

\toctitle{A Fixed-Point Model for Pancreas Segmentation in Abdominal CT Scans}
\tocauthor{Y. Zhou {\em et al.}}
\maketitle

\begin{abstract}
Deep neural networks have been widely adopted for automatic organ segmentation from abdominal CT scans.
However, the segmentation accuracy of some small organs ({\em e.g.}, the pancreas) is sometimes below satisfaction,
arguably because deep networks are easily disrupted
by the complex and variable background regions which occupies a large fraction of the input volume.
In this paper, we formulate this problem into a fixed-point model
which uses a predicted segmentation mask to shrink the input region.
This is motivated by the fact that a smaller input region often leads to more accurate segmentation.
In the training process, we use the ground-truth annotation to generate accurate input regions and optimize network weights.
On the testing stage, we fix the network parameters and update the segmentation results in an iterative manner.
We evaluate our approach on the NIH pancreas segmentation dataset,
and outperform the state-of-the-art by more than $4\%$, measured by the average Dice-S{\o}rensen Coefficient (DSC).
In addition, we report $62.43\%$ DSC in the worst case, which guarantees the reliability of our approach in clinical applications.
\end{abstract}

\section{Introduction}
\label{Introduction}

In recent years, due to the fast development of deep neural networks~\cite{Krizhevsky_2012_ImageNet}\cite{Simonyan_2015_Very},
we have witnessed rapid progress in both medical image analysis and computer-aided diagnosis (CAD).
This paper focuses on an important prerequisite of CAD~\cite{Havaei_2017_Brain}\cite{Zhou_2017_Deep},
namely, automatic segmentation of small organs ({\em e.g.}, the pancreas) from CT-scanned images.
The difficulty mainly comes from the high anatomical variability and/or the small volume of the target organs.
Indeed researchers sometimes design a specific segmentation approach
for each organ~\cite{Al-Ayyoub_2013_Brain}\cite{Roth_2015_DeepOrgan}.

Among different abdominal organs, pancreas segmentation is especially difficult,
as the target often suffers from high variability in shape, size and location~\cite{Roth_2015_DeepOrgan},
while occupying only a very small fraction ({\em e.g.}, $<0.5\%$) of the entire CT volume.
In such cases, deep neural networks can be disrupted by the background region,
which occupies a large fraction of the input volume and includes complex and variable contents.
Consequently, the segmentation result becomes inaccurate especially around the boundary areas.

To alleviate this, we apply a fixed-point model~\cite{Li_2013_Fixed} using the predicted segmentation mask to shrink the input region.
With a relatively smaller input region ({\em e.g.}, a bounding box defined by the mask),
it is straightforward to achieve more accurate segmentation.
At the training stage, we fix the input regions generated from the ground-truth annotation,
and train two deep segmentation networks, {\em i.e.}, a coarse-scaled one and a fine-scaled one,
to deal with the entire input region and the region cropped according to the bounding box, respectively.
At the testing stage, the network parameters remain unchanged,
and an iterative process is used to optimize the fixed-point model.
On a modern GPU, our approach needs around $3$ minutes to process a CT volume during the testing stage.
This is comparable to recent work~\cite{Roth_2016_Spatial}, but we report much higher accuracy.

We evaluate our approach on the NIH pancreas segmentation dataset~\cite{Roth_2015_DeepOrgan}.
Compared to recently published work~\cite{Roth_2015_DeepOrgan}\cite{Roth_2016_Spatial},
our average segmentation accuracy, measured by the Dice-S{\o}rensen Coefficient (DSC), increases from $78.01\%$ to $82.37\%$.
Meanwhile, we report $62.43\%$ DSC on the worst case,
which guarantees reasonable performance on the particularly challenging test samples.
In comparison, \cite{Roth_2016_Spatial} reports $34.11\%$ DSC on the worst case and~\cite{Roth_2015_DeepOrgan} reports $23.99\%$.
Meanwhile, our approach can be applied to segmenting other organs or tissues,
especially when the target is very small, {\em e.g.}, the pancreatic cyst~\cite{Zhou_2017_Deep}.

\section{Approach}
\label{Approach}

\subsection{Deep Segmentation Networks}
\label{Approach:Baseline}

Let a CT-scanned image be a 3D volume $\mathbf{X}$ of size $W\times H\times L$
and annotated with a ground-truth segmentation $\mathbf{Y}$ where ${y_i}={1}$ indicates a foreground voxel.
Consider a segmentation model $\mathbb{M}:{\mathbf{Z}}={\mathbf{f}\!\left(\mathbf{X};\boldsymbol{\Theta}\right)}$,
where $\boldsymbol{\Theta}$ denotes the model parameters,
and the loss function is written as $\mathcal{L}\!\left(\mathbf{Z},\mathbf{Y}\right)$.
In the context of a deep segmentation network,
we optimize $\mathcal{L}$ with respect to the network weights $\boldsymbol{\Theta}$ by gradient back-propagation.
As the foreground region is often very small,
we follow~\cite{Milletari_2016_VNet} to design a DSC-loss layer
to prevent the model from being heavily biased towards the background class.
We slightly modify the DSC of two voxel sets $\mathcal{A}$ and $\mathcal{B}$,
${\mathrm{DSC}\!\left(\mathcal{A},\mathcal{B}\right)}=
    {\frac{2\times\left|\mathcal{A}\cap\mathcal{B}\right|}{\left|\mathcal{A}\right|+\left|\mathcal{B}\right|}}$,
into a loss function between the ground-truth mask $\mathbf{Y}$ and the predicted mask $\mathbf{Z}$,
{\em i.e.}, ${\mathcal{L}\!\left(\mathbf{Z},\mathbf{Y}\right)}={1-\frac{2\times{\sum_i}z_iy_i}{{\sum_i}z_i+{\sum_i}y_i}}$.
Note that this is a ``soft'' definition of DSC,
and it is equivalent to the original form if all $z_i$'s are either $0$ or $1$.
The gradient computation is straightforward:
${\frac{\partial\mathcal{L}\!\left(\mathbf{Z},\mathbf{Y}\right)}{\partial z_j}}=
    {-2\times\frac{y_j\left({\sum_i}z_i+{\sum_i}y_i\right)-{\sum_i}z_iy_i}{\left({\sum_i}z_i+{\sum_i}y_i\right)^2}}$.

We use the 2D fully-convolutional network (FCN)~\cite{Long_2015_Fully} as our baseline.
The main reason for not using 3D models is the limited amount of training data.
To fit a 3D volume $\mathbf{X}$ into a 2D network $\mathbb{M}$, we cut it into a set of 2D slices.
This is obtained along three axes, {\em i.e.}, the coronal, sagittal and axial views.
We denote these 2D slices as $\mathbf{x}_{\mathrm{C},w}$ (${w}={1,2,\ldots,W}$),
$\mathbf{x}_{\mathrm{S},h}$ (${h}={1,2,\ldots,H}$) and $\mathbf{x}_{\mathrm{A},l}$ (${l}={1,2,\ldots,L}$),
where the subscripts $\mathrm{C}$, $\mathrm{S}$ and $\mathrm{A}$ stand for ``coronal'', ``sagittal'' and ``axial'', respectively.
We train three 2D-FCN models $\mathbb{M}_\mathrm{C}$, $\mathbb{M}_\mathrm{S}$ and $\mathbb{M}_\mathrm{A}$
to perform segmentation through three views individually (images from three views are quite different).
In testing, the segmentation results from three views are fused via majority voting.

\subsection{Fixed-Point Optimization}
\label{Approach:FixedPoint}

\newcommand{\figurewidth}{9.0cm}
\begin{figure}[t]
\begin{center}
    \includegraphics[width=\figurewidth]{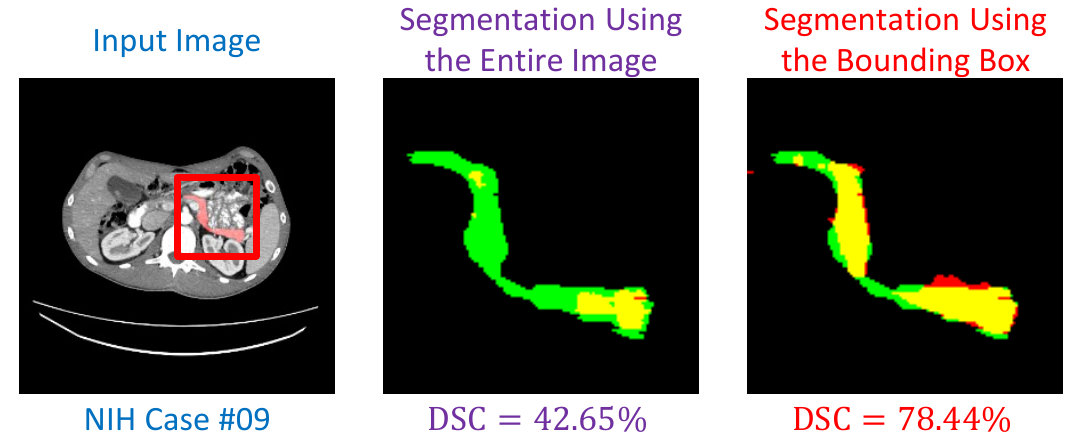}
\end{center}
\caption{
    Segmentation results with different input regions (best viewed in color),
    either using the entire image or the bounding box (the red frame).
    Red, green and yellow indicate the prediction, ground-truth and overlapped pixels, respectively.
}
\label{Fig:Comparison}
\end{figure}

The pancreas often occupies a very small part ({\em e.g.}, $<0.5\%$) of a CT volume.
It was observed~\cite{Roth_2015_DeepOrgan}
that deep segmentation networks such as FCN~\cite{Long_2015_Fully} produce less satisfying results when detecting small organs,
arguably because the network is easily disrupted by the varying contents in the background regions.
Much more accurate segmentation can be obtained by using a smaller input region around the region-of-interest.
A typical example is shown in Figure~\ref{Fig:Comparison}.

This inspires us to make use of the predicted segmentation mask to shrink the input region.
We introduce a transformation function $r\!\left(\mathbf{X},\mathbf{Z}^\star\right)$
which generates the input region given the current segmentation $\mathbf{Z}^\star$.
We rewrite the model as ${\mathbf{Z}}={\mathbf{f}\!\left(r\!\left(\mathbf{X},\mathbf{Z}^\star\right);\boldsymbol{\Theta}\right)}$,
and the loss function is
$\mathcal{L}\!\left(\mathbf{f}\!\left(r\!\left(\mathbf{X},\mathbf{Z}^\star\right);\boldsymbol{\Theta}\right),\mathbf{Y}\right)$.
Note that the segmentation mask ($\mathbf{Z}$ or $\mathbf{Z}^\star$) appears in both the input and output of
${\mathbf{Z}}={\mathbf{f}\!\left(r\!\left(\mathbf{X},\mathbf{Z}^\star\right);\boldsymbol{\Theta}\right)}$.
This is a fixed-point model, and we apply the approach described in~\cite{Li_2013_Fixed} for optimization,
{\em i.e.}, finding a steady-state solution for $\mathbf{Z}$.

\renewcommand{\figurewidth}{11.0cm}
\begin{figure}[t]
\begin{center}
    \includegraphics[width=\figurewidth]{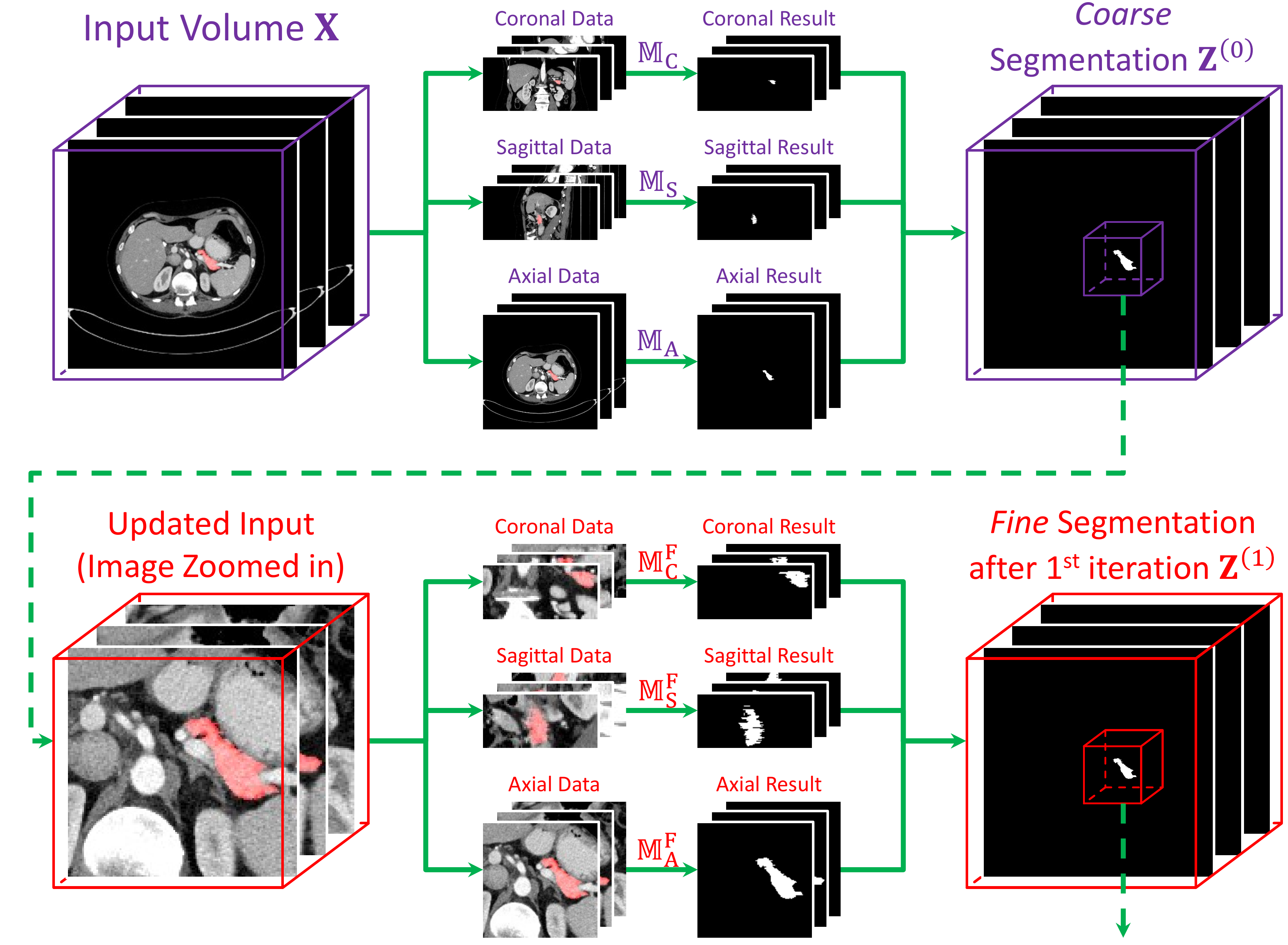}
\end{center}
\caption{
    Illustration of the testing process (best viewed in color).
    Only one iteration is shown here.
    In practice, there are at most $10$ iterations.
}
\label{Fig:Testing}
\end{figure}

{\bf In training}, the ground-truth annotation $\mathbf{Y}$ is used as the input mask $\mathbf{Z}^\star$.
We train two sets of models (each set contains three models for different views) to deal with different input sizes.
The {\em coarse-scaled} models are trained on those slices on which the pancreas occupies at least $100$ pixels
(approximately $25\mathrm{mm}^2$ in a 2D slice, our approach is not sensitive to this parameter)
so as to prevent the model from being heavily impacted by the background.
For the {\em fine-scaled} models, we crop each slice according to the minimal 2D box covering the pancreas,
add a frame around it, and fill it up with the original image data.
The top, bottom, left and right margins of the frame are random integers sampled from $\left\{0,1,\ldots,60\right\}$.
This strategy, known as data augmentation, helps to regularize the network and prevent over-fitting.

We initialize both networks using the FCN-8s model~\cite{Long_2015_Fully} pre-trained on the PascalVOC image segmentation task.
The coarse-scaled model is fine-tuned with a learning rate of $10^{-5}$ for $80\rm{,}000$ iterations,
and the fine-scaled model undergoes $60\rm{,}000$ iterations with a learning rate of $10^{-4}$.
Each mini-batch contains one training sample (a 2D image sliced from a 3D volume).

{\bf In testing}, we use an iterative process to find a steady-state solution for
${\mathbf{Z}}={\mathbf{f}\!\left(r\!\left(\mathbf{X},\mathbf{Z}^\star\right);\boldsymbol{\Theta}\right)}$.
At the beginning, $\mathbf{Z}^\star$ is initialized as the entire 3D volume,
and we compute the {\em coarse} segmentation $\mathbf{Z}^{\left(0\right)}$ using the {\em coarse-scaled} models.
In each of the following $T$ iterations, we slice the predicted mask $\mathbf{Z}^{\left(t-1\right)}$,
find the smallest 2D box to cover all predicted foreground pixels in each slice,
add a $30$-pixel-wide frame around it (this is the mean value of the random distribution used in training),
and use the {\em fine-scaled} models to compute $\mathbf{Z}^{\left(t\right)}$.
The iteration terminates when a fixed number of iterations $T$ is reached,
or the the similarity between successive segmentation results
($\mathbf{Z}^{\left(t-1\right)}$ and $\mathbf{Z}^{\left(t\right)}$) is larger than a given threshold $R$.
The similarity is defined as the inter-iteration DSC,
namely ${d^{\left(t\right)}}={\mathrm{DSC}\!\left(\mathbf{Z}^{\left(t-1\right)},\mathbf{Z}^{\left(t\right)}\right)}=
    \frac{2\times{\sum_i}z_i^{\left(t-1\right)}z_i^{\left(t\right)}}{{\sum_i}z_i^{\left(t-1\right)}+{\sum_i}z_i^{\left(t\right)}}$.
The testing stage is illustrated in Figure~\ref{Fig:Testing} and described in Algorithm~\ref{Alg:Testing}.

\begin{algorithm}[t]
\caption{Fixed-Point Model for Segmentation}
\begin{algorithmic}[1]
\State {\bf Input:} the testing volume $\mathbf{X}$,
coarse-scaled models $\mathbb{M}_\mathrm{C}$, $\mathbb{M}_\mathrm{S}$ and $\mathbb{M}_\mathrm{A}$,
fine-scaled models $\mathbb{M}_\mathrm{C}^\mathrm{F}$, $\mathbb{M}_\mathrm{S}^\mathrm{F}$ and $\mathbb{M}_\mathrm{A}^\mathrm{F}$,
threshold $R$, maximal rounds in iteration $T$.
\State {\bf Initialization:}
using $\mathbb{M}_\mathrm{C}$, $\mathbb{M}_\mathrm{S}$ and $\mathbb{M}_\mathrm{A}$
to generate $\mathbf{Z}^{\left(0\right)}$ from $\mathbf{X}$;
\For{${t}={1,2,\ldots,T}$}{}
    \State Using $\mathbb{M}_\mathrm{C}^\mathrm{F}$, $\mathbb{M}_\mathrm{S}^\mathrm{F}$ and $\mathbb{M}_\mathrm{A}^\mathrm{F}$
        to generate $\mathbf{Z}^{\left(t\right)}$ from $\mathbf{Z}^{\left(t-1\right)}$;
    \If {${\mathrm{DSC}\!\left(\mathbf{Z}^{\left(t-1\right)},\mathbf{Z}^{\left(t\right)}\right)}\geqslant{R}$}{\ {\bf break};}
    \EndIf
\EndFor
\State {\bf Output:} the final segmentation ${\mathbf{Z}^\star}={\mathbf{Z}^{\left(t\right)}}$.
\end{algorithmic}
\label{Alg:Testing}
\end{algorithm}

\section{Experiments}
\label{Experiments}

\subsection{Dataset and Evaluation}
\label{Experiments:DatasetEvalutation}

We evaluate our approach on the NIH pancreas segmentation dataset~\cite{Roth_2015_DeepOrgan},
which contains $82$ contrast-enhanced abdominal CT volumes.
The resolution of each CT scan is $512\times512\times L$,
where ${L}\in{\left[181,466\right]}$ is the number of sampling slices along the long axis of the body.
The slice thickness varies from $0.5\mathrm{mm}$--$1.0\mathrm{mm}$.
Following the standard cross-validation strategy, we split the dataset into $4$ fixed folds,
each of which contains approximately the same number of samples.
We apply cross validation, {\em i.e.}, training the model on $3$ out of $4$ subsets and testing it on the remaining one.
We measure the segmentation accuracy by computing the Dice-S{\o}rensen Coefficient (DSC) for each sample.
This is a similarity metric between the prediction voxel set $\mathcal{Z}$ and the ground-truth set $\mathcal{Y}$,
with the mathematical form of ${\mathrm{DSC}\!\left(\mathcal{Z},\mathcal{Y}\right)}=
    {\frac{2\times\left|\mathcal{Z}\cap\mathcal{Y}\right|}{\left|\mathcal{Z}\right|+\left|\mathcal{Y}\right|}}$.
We report the average DSC score together with the standard deviation over $82$ testing cases.

\subsection{Results}
\label{Experiments:Results}

We first evaluate the baseline (coarse-scaled) approach.
Using the coarse-scaled models trained from three different views
({\em i.e.}, $\mathbb{M}_\mathrm{C}$, $\mathbb{M}_\mathrm{S}$ and $\mathbb{M}_\mathrm{A}$),
we obtain $66.88\%\pm11.08\%$, $71.41\%\pm11.12\%$ and $73.08\%\pm9.60\%$ average DSC, respectively.
Fusing these three models via majority voting yields $75.74\%\pm10.47\%$,
suggesting that complementary information is captured by different views.
This is used as the starting point $\mathbf{Z}^{\left(0\right)}$ for the later iterations.

To apply the fixed-point model for segmentation,
we first compute $d^{\left(t\right)}$ to observe the convergence of the iterations.
After $10$ iterations, the average $d^{\left(t\right)}$ value over all samples is $0.9767$,
the median is $0.9794$, and the minimum is $0.9362$.
These numbers indicate that the iteration process is generally stable.

\newcommand{\colwidthA}{2.0cm}
\newcommand{\colwidthB}{1.8cm}
\begin{table}[t]
\centering
\begin{tabular}{|l||R{\colwidthA}|R{\colwidthB}|R{\colwidthB}|R{\colwidthB}|}
\hline
Method                    & Mean DSC                 & \# Iterations & Max DSC          & Min DSC          \\
\hline\hline
Roth {\em et.al}, MICCAI'2015~\cite{Roth_2015_DeepOrgan}
                          & $71.42\pm10.11$          & $-$           & $86.29$          & $23.99$          \\
\hline
Roth {\em et.al}, MICCAI'2016~\cite{Roth_2016_Spatial}
                          & $78.01\pm 8.20$          & $-$           & $88.65$          & $34.11$          \\
\hline\hline
Coarse Segmentation       & $75.74\pm10.47$          & $-$           & $88.12$          & $39.99$          \\
\hline
After $ 1$ Iteration      & $82.16\pm 6.29$          & $1$           & $\mathbf{90.85}$ & $54.39$          \\
\hline
After $ 2$ Iterations     & $82.13\pm 6.30$          & $2$           & $90.77$          & $57.05$          \\
\hline
After $ 3$ Iterations     & $82.09\pm 6.17$          & $3$           & $90.78$          & $58.39$          \\
\hline
After $ 5$ Iterations     & $82.11\pm 6.09$          & $5$           & $90.75$          & $62.40$          \\
\hline
After $10$ Iterations     & $82.25\pm 5.73$          & $10$          & $90.76$          & $61.73$          \\
\hline\hline
After ${d_t}>0.90$        & $82.13\pm 6.35$          & $1.83\pm0.47$ & $\mathbf{90.85}$ & $54.39$          \\
\hline
After ${d_t}>0.95$        & $\mathbf{82.37}\pm 5.68$ & $2.89\pm1.75$ & $\mathbf{90.85}$ & $\mathbf{62.43}$ \\
\hline
After ${d_t}>0.99$        & $82.28\pm 5.72$          & $9.87\pm0.73$ & $90.77$          & $61.94$          \\
\hline\hline
Best among All Iterations & $82.65\pm 5.47$          & $3.49\pm2.92$ & $90.85$          & $63.02$          \\
\hline
Oracle Bounding Box       & $83.18\pm 4.81$          & $-$           & $91.03$          & $65.10$          \\
\hline
\end{tabular}
\caption{
    Segmentation accuracy (measured by DSC, $\%$) reported by different approaches.
    We start from initial (coarse) segmentation $\mathbf{Z}^{\left(0\right)}$, and explore different terminating conditions,
    including a fixed number of iterations and a fixed threshold of inter-iteration DSC.
    The last two lines show two upper-bounds of our approach,
    {\em i.e.}, ``Best of All Iterations'' means that we choose the highest DSC value over $10$ iterations,
    and ``Oracle Bounding Box'' corresponds to using the ground-truth segmentation to generate the bounding box in testing.
    We also compare our results with the state-of-the-art~\cite{Roth_2015_DeepOrgan}\cite{Roth_2016_Spatial},
    demonstrating our advantage over all statistics.
}
\label{Tab:Results}
\end{table}

Now, we investigate the fixed-point model using the threshold ${R}={0.95}$ and the maximal number of iterations ${T}={10}$.
The average DSC is boosted by $6.63\%$, which is impressive given the relatively high baseline ($75.74\%$).
This verifies our hypothesis, {\em i.e.}, a fine-scaled model depicts a small organ more accurately.

We also summarize the results generated by different terminating conditions in Table~\ref{Tab:Results}.
We find that performing merely $1$ iteration is enough to significantly boost the segmentation accuracy ($+6.42\%$).
However, more iterations help to improve the accuracy of the worst case,
as for some challenging cases ({\em e.g.}, Case \#09, see Figure~\ref{Fig:Examples}),
the missing parts in coarse segmentation are recovered gradually.
The best average accuracy comes from setting ${R}={0.95}$.
Using a larger threshold ({\em e.g.}, $0.99$) does not produce accuracy gain,
but requires more iterations and, consequently, more computation at the testing stage.
In average, it takes less than $3$ iterations to reach the threshold $0.95$.
On a modern GPU, we need about $3$ minutes on each testing sample, comparable to recent work~\cite{Roth_2016_Spatial},
but we report much higher segmentation accuracy ($82.37\%$ vs. $78.01\%$).

\renewcommand{\figurewidth}{12.0cm}
\begin{figure}[t]
\begin{center}
    \includegraphics[width=\figurewidth]{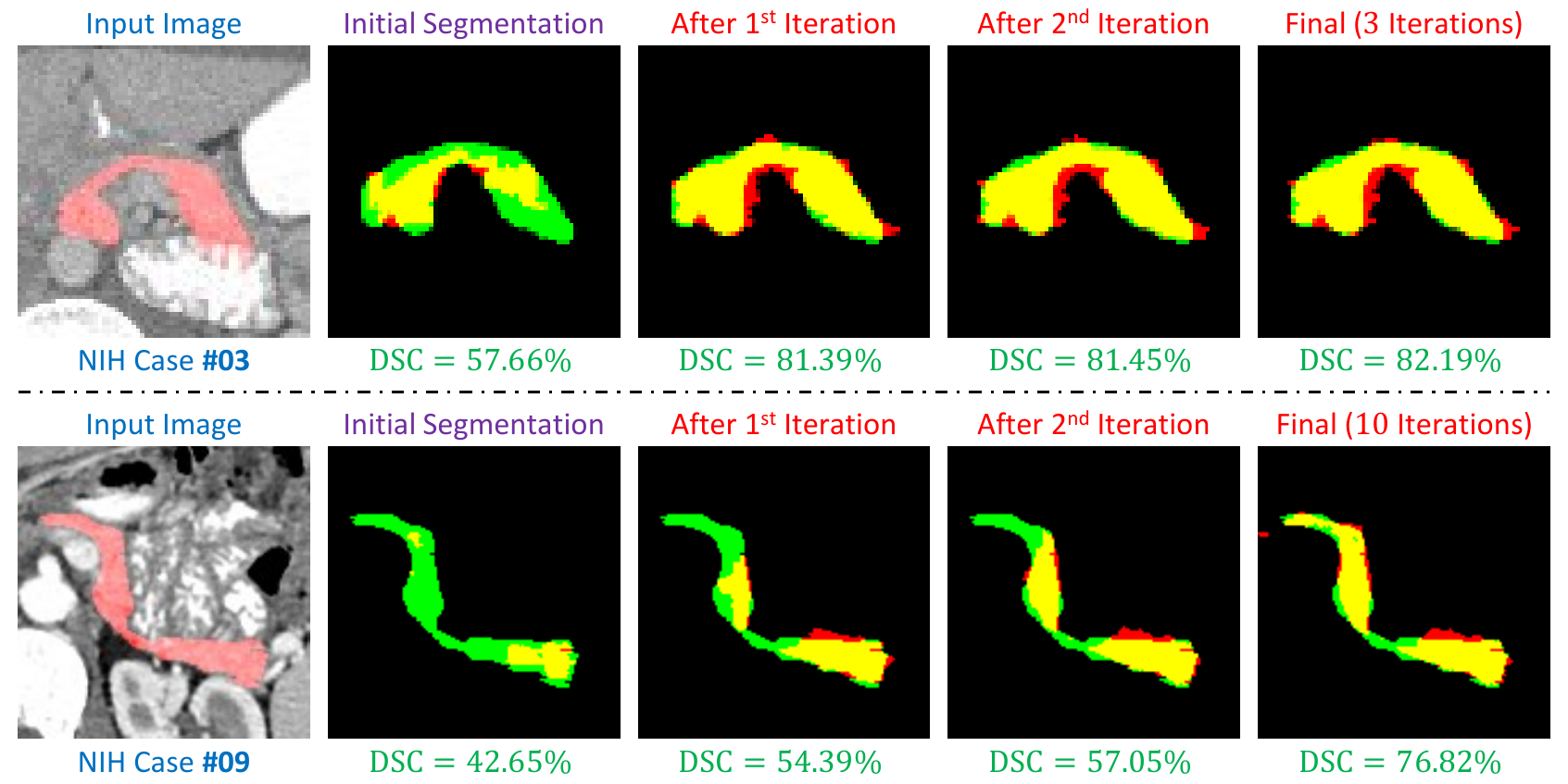}
\end{center}
\caption{
    Examples of segmentation results throughout the iteration process (best viewed in color).
    We only show a small region covering the pancreas in the axial view.
    The terminating condition is ${d^{\left(t\right)}}\geqslant{0.95}$.
    Red, green and yellow indicate the prediction, ground-truth and overlapped regions, respectively.
}
\label{Fig:Examples}
\end{figure}

As a diagnostic experiment, we use the ground-truth (oracle) bounding box of each testing case to generate the input volume.
This results in a $83.18\%$ average accuracy (no iteration is needed in this case).
By comparison, we report a comparable $82.37\%$ average accuracy,
indicating that our approach has almost reached the upper-bound of the current deep segmentation network.

We also compare our segmentation results with the state-of-the-art approaches.
Using DSC as the evaluation metric, our approach outperforms the recent published work~\cite{Roth_2016_Spatial} significantly.
The average accuracy over $82$ samples increases remarkably from $78.01\%$ to $82.37\%$,
and the standard deviation decreases from $8.20\%$ to $5.68\%$, implying that our approach are more stable.
We also implement a recently published coarse-to-fine approach~\cite{Zhang_2016_Coarse}, and get a $77.89\%$ average accuracy.
In particular, \cite{Roth_2016_Spatial} reported $34.11\%$ for the worst case
(some previous work~\cite{Chu_2013_Multi}\cite{Wang_2014_Geodesic} reported even lower numbers),
and this number is boosted considerably to $62.43\%$ by our approach.
We point out that these improvements are mainly due to the fine-tuning iterations.
Without it, the average accuracy is $75.74\%$, and the accuracy on the worst case is merely $39.99\%$.
Figure~\ref{Fig:Examples} shows examples on how the segmentation quality is improved in two challenging cases.

\section{Conclusions}
\label{Conclusions}

We present an efficient approach for accurate pancreas segmentation in abdominal CT scans.
Motivated by the significant improvement brought by small and relatively accurate input region,
we formulate a fixed-point model taking the segmentation mask as both input and output.
At the training stage, we use the ground-truth annotation to generate a smaller input region,
and train both coarse-scaled and fine-scaled models to deal with different input sizes.
At the testing stage, an iterative process is performed for optimization.
In practice, our approach often comes to an end after $2$--$3$ iterations.

We evaluate our approach on the NIH pancreas segmentation dataset with $82$ samples,
and outperform the state-of-the-art by more than $4\%$, measured by the Dice-S{\o}rensen Coefficient (DSC).
Most of the benefit comes from the first iteration,
and the remaining iterations only improve the segmentation accuracy by a little (about $0.3\%$ in average).
We believe that our algorithm can achieve an even higher accuracy if a more powerful network structure is used.
Meanwhile, our approach can be applied to other small organs,
{\em e.g.}, spleen, duodenum or a lesion area in pancreas~\cite{Zhou_2017_Deep}.
In the future, we will try to incorporate the fixed-point model into an end-to-end learning framework.

\vspace{0.3cm}
\noindent
{\bf Acknowledgements.}
This work was supported by the Lustgarten Foundation for Pancreatic Cancer Research and NSFC No. 61672336.
We thank Dr. Seyoun Park and Zhuotun Zhu for their enormous help,
and Weichao Qiu, Cihang Xie, Chenxi Liu, Siyuan Qiao and Zhishuai Zhang for instructive discussions.

\bibliographystyle{splncs03}
\bibliography{typeinst}
\end{document}